\documentclass[letterpaper, 10 pt, conference]{ieeeconf}  

\IEEEoverridecommandlockouts                              

\usepackage{times}
\usepackage{epsfig}
\usepackage{graphicx}
\usepackage{amsmath}
\usepackage{amssymb}
\usepackage{xspace}
\usepackage{float}

\usepackage{booktabs}
\usepackage{mathtools}
\usepackage{xcolor}

\usepackage{amsfonts}

\usepackage{epstopdf}
\usepackage{color}

\usepackage{multirow}
\usepackage{pbox}

\usepackage{amsopn}
\usepackage{caption}
\usepackage{graphicx}
\usepackage{subcaption}
\usepackage{multirow}
\usepackage{longtable}
\usepackage{array}
\usepackage{makecell}

\usepackage{tikz}
\usepackage[table]{xcolor}

\usepackage{hyperref}

\DeclareRobustCommand\onedot{\futurelet\@let@token\@onedot}
\def\@onedot{\ifx\@let@token.\else.\null\fi\xspace}

\newcommand{\tableline}{\noalign{\hrule height 1.5pt}} 

\definecolor{ubpubColor}{rgb}{0.43, 0.5, 0.5}

\newcommand{\subsubsubsection}[1]{
	\par\vspace{0.0cm}\textbf{#1:}\quad
	\ignorespaces
}
\newcommand{\CP}[1]{\ignorespaces}

\newcolumntype{C}[1]{>{\centering\arraybackslash}m{#1}} %
\newcolumntype{L}[1]{>{\raggedright\arraybackslash}m{#1}}
\title{\LARGE \bf
Seg2Track-SAM2: SAM2-based Multi-object Tracking and Segmentation
}

\author{D. Mendonça, T. Barros,  C. Premebida, U.J. Nunes 
\thanks{Authors are with the University of Coimbra, Institute of Systems and Robotics, Department of Electrical and Computer Engineering, Portugal. Emails: \footnotesize\{diogo.mendonca,~tiagobarros,~urbano,~cpremebida\}@isr.uc.pt.}
}

\begin{document}

\maketitle
\thispagestyle{empty}
\pagestyle{empty}

\begin{abstract}
Autonomous-driving perception systems require robust Multi-Object Tracking (MOT) to operate reliably in dynamic environments. MOT maintains consistent object identities across frames while preserving spatial accuracy. Recent foundation models, such as SAM2, provide promptable video segmentation without task-specific fine-tuning. However, their direct application to Multi-Object Tracking and Segmentation (MOTS) remains limited by the absence of explicit identity management mechanisms and by growing memory requirements during tracking.
This work introduces Seg2Track-SAM2, a framework that integrates pretrained object detectors with SAM2 and a dedicated Seg2Track module to support track initialization, data association, and track refinement. The method operates without dataset-specific fine-tuning and remains detector-agnostic.
Experimental evaluation on the KITTI MOTS and MOTS Challenge benchmarks shows that Seg2Track-SAM2 ranks fourth overall in both datasets while achieving the highest association accuracy (AssA) among compared methods. In addition, a sliding-window memory strategy reduces memory usage by up to 75\% with minimal impact on tracking performance, enabling deployment under resource constraints.
Together, these results indicate that Seg2Track-SAM2 improves identity consistency and memory efficiency in MOTS without requiring dataset-specific training.
The code is available at \url{https://github.com/hcmr-lab/Seg2Track-SAM2}
\end{abstract}


\begin{figure}[!t]
\includegraphics[width=0.5\textwidth, trim={0.55cm 0.5cm 0.1cm 0.3cm},clip]{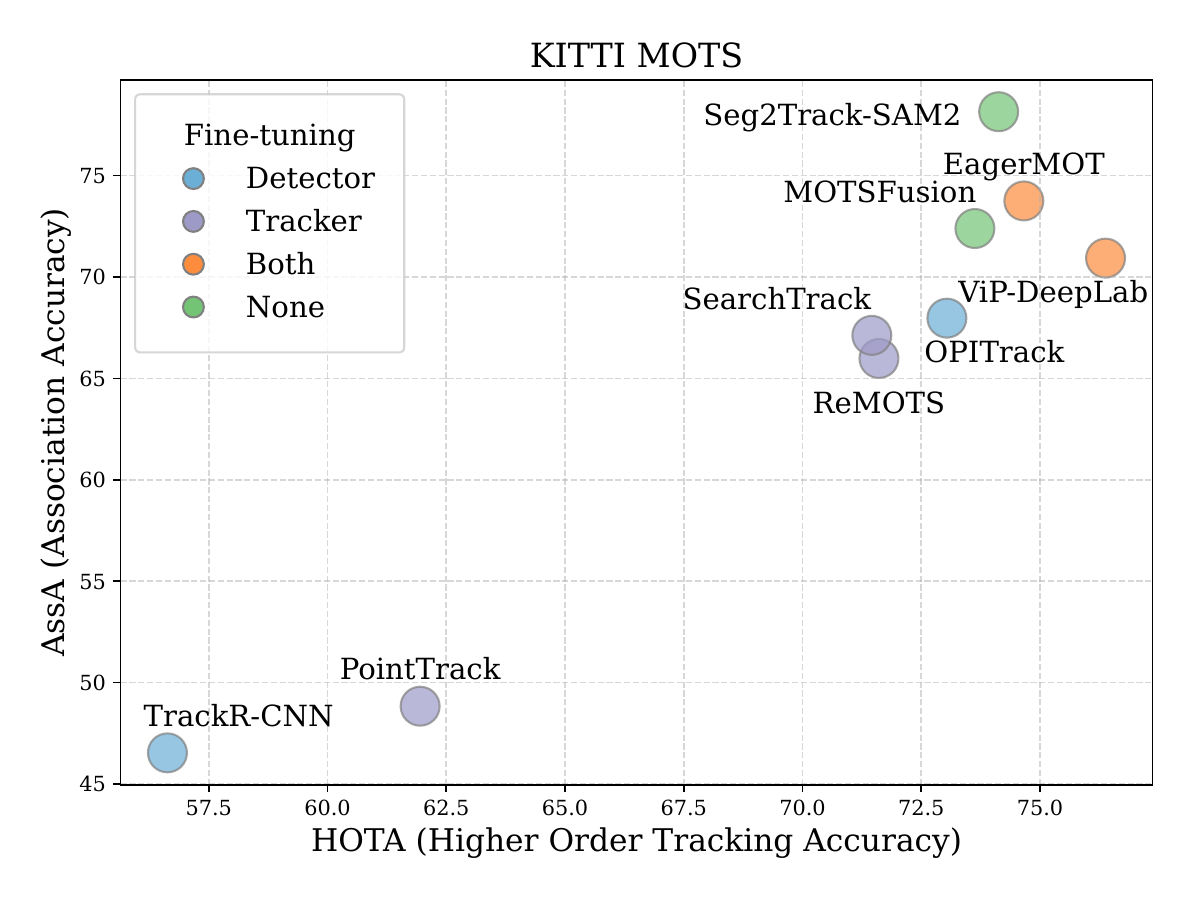}
\vspace{10pt}
\includegraphics[width=0.5\textwidth, trim={0.55cm 0.5cm 0.1cm 0.3cm},clip]{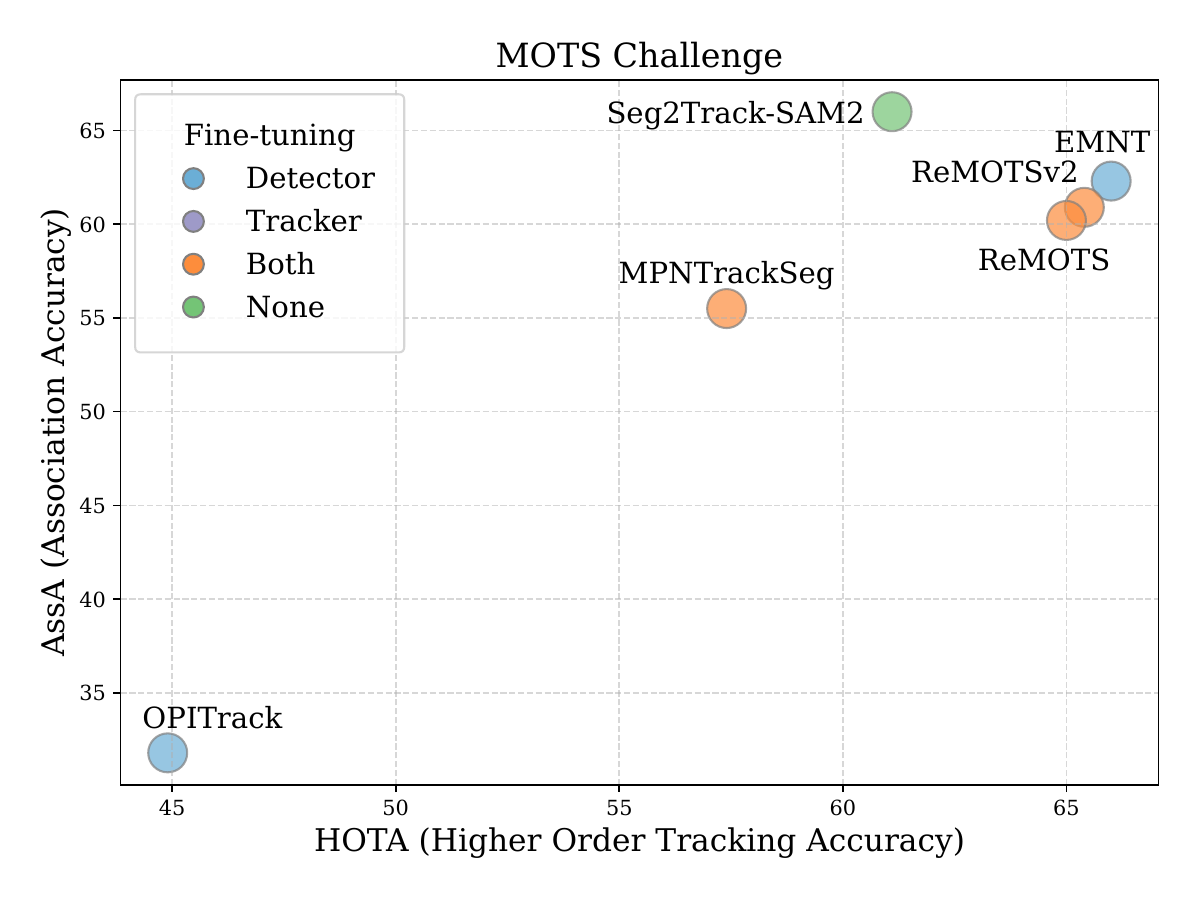}
    \caption{Illustration of a comparative analysis of MOTS methods on the KITTI MOTS (top) and MOTS Challenge (below) benchmarks. The KITTI MOTS graph shows each method's performance based on its average HOTA against AssA for the car and pedestrian classes. The data points are color-coded to indicate the training dependency of each approach on a detector, a tracker, both, or neither. The figure shows that Seg2Track-SAM2 achieved the highest AssA score and is ranked in the top four for overall HOTA performance on both datasets. The proposed approach does not require fine-tuning on either a detector or a tracker for the target datasets.}
    \label{fig:method_comparison}
\end{figure}
\section{INTRODUCTION}


Autonomous-driving perception systems rely on Multi-Object Tracking (MOT) to maintain consistent object identities across image sequences, enabling trajectory estimation and decision-making in dynamic environments. Identity preservation becomes particularly challenging under occlusion, appearance variation, and complex motion patterns.


Multi-Object Tracking and Segmentation (MOTS) extends MOT by incorporating pixel-level masks, requiring simultaneous detection, segmentation, and temporal association. This joint formulation increases sensitivity to identity switches and false-positive propagation, as both spatial delineation and temporal consistency must be maintained.

Recent foundation models have been applied to video segmentation tasks. The Segment Anything Model 2 (SAM2)~\cite{ravi2024sam} provides promptable video segmentation without task-specific fine-tuning. However, its direct application to MOTS remains limited by the absence of explicit identity management and by unbounded memory growth during tracking~\cite{jiang2025sam2mot}. In particular, SAM2 does not include dedicated mechanisms for track initialization, association refinement, or long-term identity consistency.


This work introduces Seg2Track-SAM2, a framework that integrates pretrained object detectors with SAM2 and a dedicated tracking module for class-aware MOTS without dataset-specific fine-tuning. The Seg2Track module performs track initialization, association, and refinement through two components: \textit{Track Quality Assessment} (TQA), which evaluates temporal consistency and filters unreliable tracks, and \textit{Object Association and Filtering} (OAF), which reinforces ambiguous tracks and limits false-positive propagation. A sliding-window memory strategy further constrains memory usage during tracking.

The experimental evaluation was conducted on the KITTI Multi-Object Tracking and Segmentation (KITTI MOTS) and Multi-Object Tracking and Segmentation Challenge (MOTS Challenge) benchmarks ~\cite{voigtlaender2019mots}.  Without dataset-specific fine-tuning, Seg2Track-SAM2 ranks fourth overall on both datasets and achieves the highest association accuracy (AssA),as outlined in Fig.~\ref{fig:method_comparison}. 



The main contributions of this work are summarized as follows:
\begin{itemize}
    \item A detector-agnostic MOTS framework, Seg2Track-SAM2, that incorporates persistent tracking mechanisms into SAM2 through Track Quality Assessment and Object Association and Filtering modules. These modules enhance temporal consistency for active tracks, reduce the propagation of false positives, and preserve consistent object identities across frames.
    \item A evaluation on KITTI MOTS and MOTS Challenge without dataset-specific fine-tuning.
    \item A sliding-window memory strategy that reduces memory usage while maintaining tracking accuracy.
    \item Establishment of a new state-of-the-art result in association accuracy (AssA) on both benchmarks.
\end{itemize}

\begin{figure*}[ht]
    \centering
    \includegraphics[width=\textwidth, trim={0cm 0cm 0cm 0.0cm},clip]{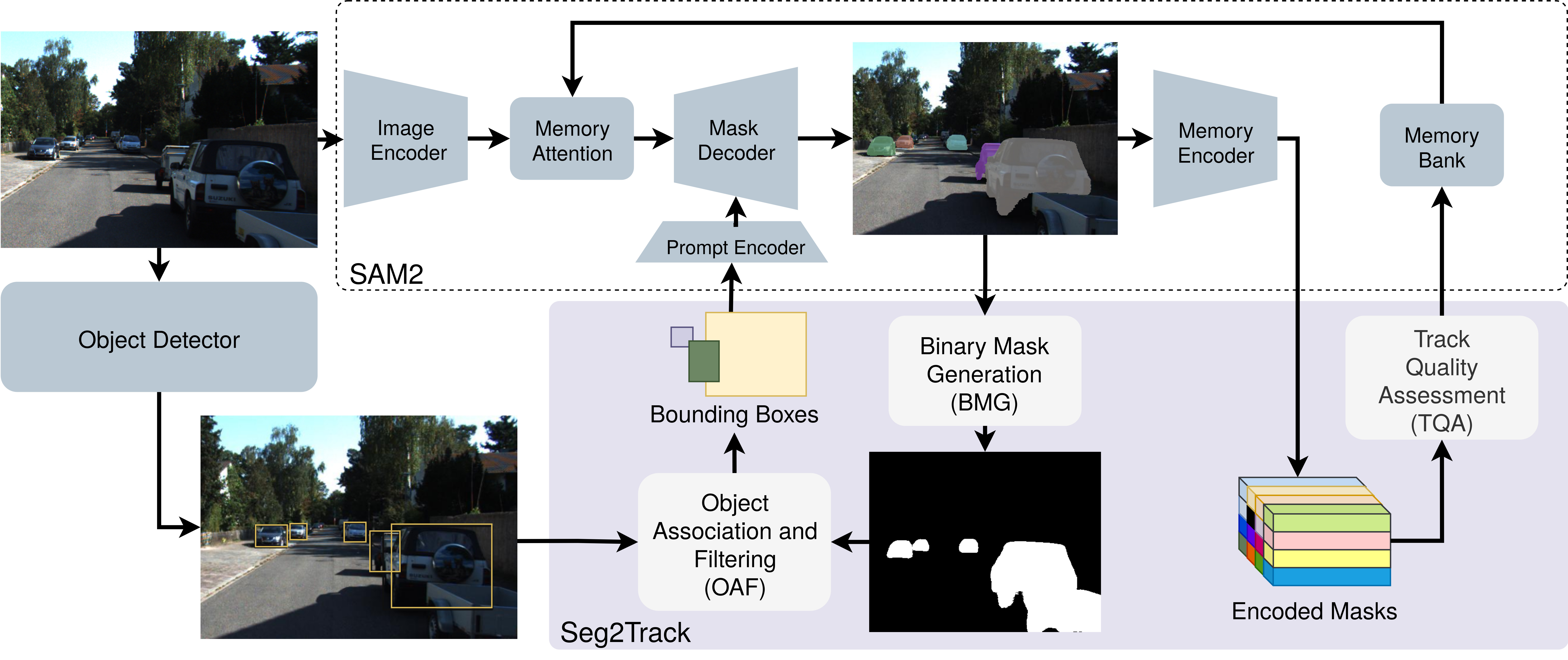}
    \caption{Illustration of the Seg2Track-SAM2 approach. The Seg2Track-SAM2 approach is a comprehensive system that combines an object detector, SAM2, and a track management module (Seg2Track) to perform robust MOTS. An input image is simultaneously analyzed by the object detector to generate bounding box proposals and by SAM2 to create segmentation masks. The Seg2Track module then uses these bounding boxes and previously generated masks to manage object tracks over time, initiating new tracks, reinforcing existing tracks, and removing stale ones through a series of association, filtering, and quality assessment processes. }
    \label{fig:pipeline}
\end{figure*}

\section{RELATED WORK}

MOT approaches are commonly categorized into two families: \textit{Tracking-by-Detection} (TBD) and \textit{Joint Detection and Tracking} (JDT). JDT methods integrate detection and association within a single deep learning framework, simultaneously optimizing object detection and re-identification (ReID), as exemplified by FairMOT~\cite{zhang2021fairmot}. In contrast, TBD methods separate the detection and data association stages, typically applying an object detector first and then linking detections across frames~\cite{zhang2021fairmot,10.1007/978-3-031-19812-0_38}. In this work, we address the problem from the TBD perspective, focusing specifically on the MOTS task.

Early TBD approaches such as SORT~\cite{7533003} employ a Kalman filter~\cite{welch1995introduction} to predict object locations and the Hungarian algorithm~\cite{kuhn1955hungarian} for efficient IoU-based association. DeepSORT~\cite{8296962} extends this framework by incorporating deep ReID embeddings, improving robustness under occlusion. ByteTrack~\cite{zhang2022bytetrack} further advances association by considering both high- and low-confidence detections, enabling the tracker to follow almost every object instance. More recently, SearchTrack~\cite{tsai2022searchtrack} addresses association through object-customized search and motion-aware features, improving discriminability in crowded scenes.

Beyond holistic object features, several works have explored a \textit{tracking-by-points} paradigm. PointTrack~\cite{xu2020Segment} samples discriminative pixels from instance masks and learns instance embeddings directly from these points, treated as an unordered set, for robust association. Similarly, OPITrack~\cite{9881968} formulates tracking as a point-set matching problem, leveraging segmentation masks to enhance robustness in MOTS. In contrast, ReMOTS~\cite{yang2020remots} employs a recurrent mask refinement network that iteratively improves both segmentation and association, using a self-supervised scheme based on predicted masks and tracklets.

More recent research works have incorporated 3D information to enhance MOT. EagerMOT~\cite{9562072} integrates camera and LiDAR detections for improved spatio-temporal modeling, while ViP-DeepLab~\cite{Qiao_2021_CVPR} performs depth-aware video panoptic segmentation, combining monocular depth estimation with segmentation to support 3D multi-object tracking.

In vision-based tracking, foundation models have influenced recent approaches, including SAM2~\cite{ravi2024sam}, which provides promptable video segmentation. Trained on large-scale datasets, SAM2 can segment and track objects without task-specific fine-tuning. However, it does not include dedicated mechanisms for persistent identity management in MOTS settings.

The proposed framework builds upon SAM2 by integrating pretrained object detectors to generate initial object hypotheses, using SAM2 for mask propagation, and introducing the Seg2Track module to perform track initialization, data association, and track refinement. By incorporating explicit identity management and track control mechanisms, the framework addresses limitations of SAM2 in maintaining consistent object identities over time.

\section{METHODOLOGY}
The proposed framework addresses MOTS, which can be defined as detecting, segmenting, and tracking multiple objects across a sequence of $T$ image frames $\{I_1, I_2, \dots, I_T\}$, where each frame $I_t \in \mathbb{R}^{H \times W \times C}$ has height $H$, width $W$, and $C$ channels. The goal is to predict $K$ object trajectories $\mathcal{T} = \{T_1, T_2, \dots, T_K\}$, where each trajectory $T_k = \{(t, m_{t,k}, s_{t,k}, c_{t,k},\text{id}_k)\}_{t \in F_k}$ includes detections for object $k$ with identity $\text{id}_k \in \{1, 2, \dots, K\}$. Here, $F_k \subseteq \{1, 2, \dots, T\}$ denotes frames where object $k$ appears, $m_{t,k} \in \{0,1\}^{H \times W}$ is the binary segmentation mask, and $s_{t,k} \in [0,1]$ is the confidence score of the object belonging to class $c_{t,k} \in \{1,2,\dots, L\}$.  The proposed approach, outlined in Fig.~\ref{fig:pipeline}, combines a detector generating bounding box proposals, SAM2 for segmentation and tracking, and a Seg2Track module for track and object management.

\subsection{Seg2Track-SAM2}
Specifically, the proposed framework integrates three components: (i) object detector, which generates object proposals for each input image; (ii) SAM2 that is responsible for tracking; and the (iii) Seg2Track that performs track initialization, data association, and track refinement based on new object proposals and previous object tracks.

\subsubsection{Object Proposals}

 We use a SOTA detector to generate bounding box proposals for each frame. Formally, the detector is a function $\phi_{\text{det}}: \mathbb{R}^{H \times W \times C} \to \mathbb{D}$, where $\mathbb{D}= \{\{b_i\}_{i=1}^{N_d} \mid b_i \in \mathbb{R}^4\} $ defines the output space of the detector and $b_i$ is a bounding box. For frame $I_t$, the detector outputs 
 \begin{equation}
    D_t = \phi_{\text{det}}(I_t) = \{b_{t,1}, b_{t,2}, \dots, b_{t,N_t}\} \in \mathbb{D}.  
 \end{equation}
 These bounding box proposals drive SAM2 and are processed by Seg2Track to initialize or update tracks. Relying on an object detector to prompt SAM2 allows the framework to handle different object classes, with Seg2Track assigning each object the class\footnote{Notice that SAM2 has no class-awareness.} of its initializing detection.
 

\subsubsection{Segment Anything Model 2} SAM2 is a foundation model for segmentation-based tracking with robust generalization~\cite{ravi2024sam}, and is used for tracking in the Seg2Track-SAM2 framework. SAM2's output is defined as $\phi_{\text{sam}}: \mathbb{R}^{H \times W \times C} \times \mathbb{D} \to \mathbb{M}$, with
\begin{align*}
\mathbb{M} = \Bigl\{ \bigl\{ (m_i, s_i, e_i,\text{id}_i) \bigr\}_{i=1}^{N_M} \mid & N_M \geq 0, \\
& m_i \in [0,1]^{H \times W}, \\
& s_i \in [0,1], \\
& e_i \in \mathbb{R}^{H_E \times W_E}, \\
& \text{id}_i \in  \{1, 2, . . . , K\} \Bigr\},
\end{align*}
where each tuple $(m_i, s_i, e_i, \text{id}_i)\in \mathbb{M}$ includes a soft segmentation mask $m_i$, IoU-based confidence score $s_i$~\cite{ravi2024sam}, encoded mask $e_i$ (downsampled via SAM2’s memory encoder), and identity $\text{id}_i$. 
Hence, for each frame $I_t$ and the corresponding bounding box prompts $P_t = \{p_{t,i}\}_{i=1}^{N_P} \in \mathbb{D}$, SAM2 outputs a set of output vectors $M_t = \{(m_{t,i}, s_{t,i}, e_{t,i}, \text{id}_{t,i})\}_{i=1}^{N_t} \in \mathbb{M}$, which is used in Seg2Track.

Additionally, we modify SAM2's Memory Bank to enhance memory efficiency during tracking. To do so, we adopt a sliding window-based approach that limits the number of past states considered when computing the mask for frame $t$. In contrast, the original implementation uses all preceding states up to $t-1$, resulting in unbounded memory growth and increased memory consumption.
In the proposed sliding window modification, we only retain the most recent $T_w$ states. This sliding window approach bounds the memory footprint by constraining past states, which ensures a consistent computational load across frames. By balancing efficiency with the retention of essential temporal information, the method enables scalable deployment on memory-constrained systems while maintaining reliable track association and mask refinement.

\subsubsection{Seg2Track}
Seg2Track module manages track initialization, track management, and refinement. Seg2Track integrates three key blocks: Track Quality Assessment, which evaluates the quality of predicted masks to decide whether to maintain, remove, or reinforce a track; Binary Mask Generation, which merges individual object masks from the previous frame into a single binary mask; and Object Association and Filtering, a two-stage process that uses incoming object proposals and track quality assessments to determine if a new track should be initiated or if an existing track should be reinforced by reprompting SAM2.

\subsubsubsection{Track Quality Assessment}\label{sec:tqa} 
This module evaluates mask quality in frame $t$ using the IoU-based confidence score provided by SAM2, classifying masks into \textit{High}, \textit{Uncertain}, or \textit{Low} states based on thresholds $\tau_h$ and $\tau_l$. Let $M_t$ be the set of output vectors generated by SAM2. Each vector $(m_{t,i}, s_{t,i}, e_{t,i}, \text{id}_{t,i}) \in M_t$ is assigned a state according to the following criteria:
\begin{equation}
\label{equ:States}
S_{t,i} =
\begin{cases}
\textit{High}, & s_{t,i} > \tau_h, \\
\textit{Uncertain}, & \tau_l < s_{t,i} \leq \tau_h, \\
\textit{Low}, & s_{t,i} \leq \tau_l,
\end{cases}
\end{equation}
For masks classified as \textit{High}, the corresponding embedding $e_{t,i}$ is used to update SAM2's memory bank. If a mask is classified as \textit{Uncertain}, it is passed to the two-stage matching process without updating the memory bank. For a \textit{Low} state, a counter for consecutive Low classifications is incremented, and the track is removed after $n_{tries}$ occurrences.

\subsubsubsection{Binary Mask Generation}
The proposed Seg2Track module receives SAM2's output vectors $M_{t}$, and merges the individual object masks $m_{t,i} \in M_{t}$, into a single binary mask $N_{t} \in \{0,1\}^{H \times W}$. Hence, the unified mask $N_{t}$ is given by:
\begin{equation}
N_{t}(x, y) = \bigvee_{i=1}^{N_{t}} m_{t,i}(x, y)
\end{equation}
This mask is used in the matching process to identify overlaps with new proposals.

\subsubsubsection{Object Association and Filtering}
This stage is responsible for both initializing new tracks and reinforcing existing ones. Its objective is to determine which bounding boxes should be used to prompt SAM2. To this end, it relies on two inputs: (i) a set of object proposals, $D_t = \{b_{t,j}\}_{j=1}^{N_d}$, detected in frame $I_t$; and (ii) SAM2’s output vectors from the previous frame, $M_{t-1} = \{(m_{t-1,i}, s_{t-1,i}, e_{t-1,i}, \text{id}_{t-1,i})\}_{i=1}^{N_{t-1}}$, corresponding to frame $I_{t-1}$. The output of this stage is a subset of bounding boxes, $P_t \subseteq D_t$, which are subsequently used to prompt SAM2.

Proposals likely corresponding to existing tracks are identified by computing the Intersection-over-Union (IoU) between each bounding box $b_{t,j} \in D_t$ and the union of masks from the previous frame, denoted $N_{t-1}$:
\begin{equation}
\label{equ:IoU}
v_j = \frac{|b_{t,j} \cap N_{t-1}|}{|b_{t,j} \cup N_{t-1}|}, 
\quad \forall b_{t,j} \in D_t.
\end{equation}
Bounding boxes with $v_j \geq \tau_v$ are retained as candidate matches, forming the subset
\begin{equation}
D'_t = \{b_{t,j} \mid v_j \geq \tau_v\} \subseteq D_t.
\end{equation}
Conversely, proposals with $v_j < \tau_v$ are treated as potential new objects and are used to initialize tracks with unique identities. This filtering step leverages the binary mask $N_{t-1}$ to handle occlusions, ensuring that proposals with significant overlaps are prioritized. Since SAM2 does not differentiate object categories, Seg2Track assigns to each new track the class label of the detection used for initialization.


To reinforce existing tracks, proposals in $D'_t$ are associated with masks from the previous frame using the Hungarian algorithm~\cite{kuhn1955hungarian}. The cost metric is the Euclidean distance between the bounding-box centers and the mask centers. This results in a set of associations
\begin{equation}
A_t = \{(m_{t-1,i}, b'_{t,j}) \}.
\end{equation}
We retain only those pairs where the track $m_{t-1,i}$ is in the \emph{Uncertain} state, as determined by the Track Quality Assessment. This ensures that only ambiguous tracks are updated. The final set of bounding boxes used to prompt SAM2 is defined as follows:
\begin{align}
P_t = \{b'_{t,j} \mid  \quad & (m_{t-1,i}, b'_{t,j}) \in A_t, \; \tau_l < s_{t-1,i} \leq \tau_h \}\\
 & \;\cup\; \{b_{t,j} \mid v_j < \tau_v, \quad \forall b_{t,j} \in D_t\}, \nonumber
\end{align}
which includes both proposals reinforcing uncertain tracks and proposals used for initializing new tracks.

The threshold values $\tau_h$, $\tau_l$ in (\ref{equ:States}), and the value $\tau_v$ for the object association proposals (\ref{equ:IoU})  were obtained empirically, and are discussed in Section~\ref{sec:implementation}.

\begin{table*}[thp]
\centering
\caption{Results on the 2D KITTI MOTS benchmark (for \textit{car} and \textit{pedestrian} classes). The superscript (in blue) indicates the column-wise ranking of the methods.}
 {\renewcommand{\arraystretch}{1.2}
\begin{tabular}{l|cccc|cccc}
\tableline
 & \multicolumn{4}{c|}{Car} & \multicolumn{4}{c}{Pedestrian} \\
 Method & HOTA $\uparrow$ & DetA $\uparrow$ & AssA $\uparrow$&  LocA $\uparrow$& HOTA $\uparrow$ & DetA $\uparrow$ & AssA $\uparrow$ &  LocA $\uparrow$\\
\hline
 ViP-DeepLab~\cite{Qiao_2021_CVPR} & 76.38{\color{blue}$^1$} & 82.70{\color{blue}$^1$}  & 70.93  & 90.75{\color{blue}$^1$}  & 64.31{\color{blue}$^1$} & 70.69{\color{blue}$^1$} & 59.48{\color{blue}$^3$} &  84.40{\color{blue}$^1$} \\
 EagerMOT~\cite{9562072} & 74.66{\color{blue}$^2$} & 76.11 & 73.75{\color{blue}$^2$} & 90.46{\color{blue}$^2$} & 57.65& 60.30 & 56.19 & 83.65 \\
OPITrack~\cite{gullapalli2021opitrack} & 73.04 & 79.44{\color{blue}$^2$}  & 67.97 & 88.57 & 60.38{\color{blue}$^2$} & 62.45 & 60.05{\color{blue}$^2$} & 83.55 \\
 ReMOTS~\cite{yang2020remots} & 71.61  & 78.32 & 65.98  &  89.33 & 58.81 & 67.96{\color{blue}$^2$} & 52.38 & 84.18{\color{blue}$^2$}  \\ 
 SearchTrack~\cite{tsai2022searchtrack} & 71.46  & 76.76 & 67.12 &  88.08 & 57.63  & 63.66{\color{blue}$^3$} & 53.12 & 80.89\\
 MOTSFusion~\cite{luiten19arxiv}  & 73.63 & 75.44 & 72.39{\color{blue}$^3$} & 90.29{\color{blue}$^3$} & 54.04 & 60.83 & 49.45 & 83.71{\color{blue}$^3$}  \\
 PointTrack~\cite{xu2020Segment}  & 61.95 & 79.38{\color{blue}$^3$}  & 48.83 & 88.52 & 54.44 & 62.29 & 48.08 & 83.28\\
 TrackR-CNN~\cite{voigtlaender2019mots} & 56.63 & 69.90 & 46.53 &   86.60  & 41.93 & 53.75 & 33.84 & 78.03 \\  \hline
 \textit{(No fine-tuning w/YOLOv11)} &  &  & &    &  &  &  &  \\
 Baseline (SAM2) & 67.53 & 64.31 & 71.60 & 87.39 & 51.12 & 50.10 & 57.56 & 80.18\\
 Ours (Seg2Track-SAM2)  & 74.13{\color{blue}$^3$}  & 71.03{\color{blue}$^8$} & \textbf{78.15}{\color{blue}$^1$} & 89.69{\color{blue}$^4$}  & 60.00{\color{blue}$^3$} & 56.61{\color{blue}$^8$} & \textbf{65.86}{\color{blue}$^1$} & 80.40{\color{blue}$^8$} \\ 
 \tableline
\end{tabular}
}
\label{tab:mots}
\end{table*}

\section{EXPERIMENTAL EVALUATION}

This section presents the evaluation framework, detailing the dataset benchmarks, implementation details, empirical results and discussion.

\subsection{Benchmarks} 
The experimental evaluation of the proposed approach is conducted on two widely used benchmarks for multi-object tracking and segmentation: the KITTI MOTS dataset~\cite{Geiger12CVPR} and the MOTS Challenge~\cite{milan2016mot16benchmarkmultiobjecttracking}.

\subsubsection{KITTI MOTS} 
The KITTI Multi-Object Tracking and Segmentation (MOTS) benchmark extends the original KITTI tracking dataset ~\cite{Geiger12CVPR} by providing dense, pixel-level segmentation annotations for each object instance. The dataset comprises 21 training sequences and 29 test sequences captured from a moving platform in urban driving scenarios. 

Annotations are provided for the object classes \textit{Car} and \textit{Pedestrian}. Unlike bounding-box-only tracking benchmarks, KITTI MOTS evaluates the joint task of detection, segmentation, and temporal association, requiring methods to produce consistent instance masks over time.

\subsubsection{MOTS Challenge} 
The MOTS Challenge generalizes the MOTS task to more diverse and crowded scenarios. It includes sequences derived from the MOTChallenge 2017 benchmark~\cite{milan2016mot16benchmarkmultiobjecttracking}, and provides pixel-accurate instance segmentation annotations for tracked objects. In particular, the MOTS subset focuses on crowded pedestrian scenes, introducing significant challenges such as frequent occlusions, scale variations, and dense interactions. Annotations are only provided for the \textit{Pedestrian} class. 

\subsubsection{Evaluation Metrics} Evaluation on both the KITTI MOTS and MOTS Challenge datasets follows the official benchmark protocol, with methods ranked primarily by the Higher Order Tracking Accuracy (HOTA) metric~\cite{Luiten2020IJCV}. Unlike earlier measures such as MOTA or MOTP, HOTA jointly evaluates detection (DetA), association (AssA), and localization (LocA), providing a balanced evaluation of tracking performance. It should be noted that these metrics are computed over varying detection thresholds and subsequently averaged, reducing bias toward either precision or recall and providing a more robust indicator of overall tracking quality.

\begin{table*}[thp]
\centering
\setlength{\tabcolsep}{4pt}
\caption{Comparison of tracking performance across KITTI MOTS (Car and Pedestrian classes) and MOTS Challenge training sets using multiple pretrained detectors with Seg2Track-SAM2.}
{\renewcommand{\arraystretch}{1.5}
\begin{tabular}{l|ccc|ccc|ccc|ccc}
\tableline
 & \multicolumn{3}{c|}{KITTI MOTS (Car)} 
 & \multicolumn{3}{c|}{KITTI MOTS (Pedestrian)} 
 & \multicolumn{3}{c|}{MOTS Challenge}
 & \multicolumn{3}{c}{Average} \\
Method & HOTA $\uparrow$ & DetA $\uparrow$ & AssA $\uparrow$
& HOTA $\uparrow$ & DetA $\uparrow$ & AssA $\uparrow$
& HOTA $\uparrow$ & DetA $\uparrow$ & AssA $\uparrow$
& HOTA $\uparrow$ & DetA $\uparrow$ & AssA $\uparrow$\\
\hline

Ours w/RT\_DETR ~\cite{10657220}
& 73.309{\color{blue}$^2$} & 68.288{\color{blue}$^3$} & 79.078{\color{blue}$^1$}
& 48.113{\color{blue}$^3$} & 37.291{\color{blue}$^3$} & 62.882{\color{blue}$^3$}
& 62.764{\color{blue}$^3$} & 62.160{\color{blue}$^3$} & 64.818{\color{blue}$^3$}
& 59.062{\color{blue}$^3$} & 55.913{\color{blue}$^3$} & 68.926{\color{blue}$^3$}\\

Ours w/YOLOv11 ~\cite{khanam2024yolov11overviewkeyarchitectural}
& 74.632{\color{blue}$^1$} & 70.896{\color{blue}$^1$} & 78.934{\color{blue}$^2$}
& 51.731{\color{blue}$^1$} & 41.532{\color{blue}$^1$} & 65.191{\color{blue}$^1$}
& 64.093{\color{blue}$^2$} & 63.571{\color{blue}$^2$} & 65.880{\color{blue}$^1$} 
& 63.485{\color{blue}$^1$} & 58.666{\color{blue}$^1$} & 70.002{\color{blue}$^1$} \\

Ours w/YOLOv26 \cite{sapkota2026yolo26keyarchitecturalenhancements}
& 73.348{\color{blue}$^3$} & 68.720{\color{blue}$^2$} & 78.659{\color{blue}$^3$}
& 49.974{\color{blue}$^2$} & 39.299{\color{blue}$^2$} & 64.290{\color{blue}$^2$}
& 65.017{\color{blue}$^1$} & 66.064{\color{blue}$^1$} & 65.313{\color{blue}$^2$}
& 62.780{\color{blue}$^2$} & 58.028{\color{blue}$^2$} & 69.421{\color{blue}$^2$}\\

\tableline
\end{tabular}
}
\label{tab:detector_comparison}
\end{table*}

\subsection{Implementation Details}
\label{sec:implementation}

This section elaborates on the implementation details to generate the reported  results.    
Seg2Track-SAM2 operates without dataset-specific fine-tuning. For the detection stage, we employ three pretrained detectors.
For the detection stage, we employ three pretrained detectors (YOLOv11 ~\cite{khanam2024yolov11overviewkeyarchitectural}, YOLOv26 \cite{sapkota2026yolo26keyarchitecturalenhancements} and RT\_DETR ~\cite{10657220}), which are applied directly to the KITTI MOTS and MOTS Challenge sequences.
Segmentation relies on SAM2.1-large, whose pre-trained weights provide reliable mask generation.

In order to minimize false positive prediction, only detection proposals with a confidence score above 0.50 forwarded to the Seg2Track module. Within the Track Quality Assessment module, mask quality thresholds are fixed at $\tau_h=0.70$ and $\tau_l=0.10$, with tracks removed only after $n_{tries}=5$ consecutive low-confidence states. Object Association and Filtering module applies class-specific thresholds for $\tau_v = 0.6$ for cars and 0.85 for pedestrians, to accommodate differences in object scale and susceptibility to occlusion. 

To further ensure computational efficiency, we limit the number of past states considered by SAM2 during mask computation. Unless otherwise specified, all experiments reported in this work utilize a $T_w$ of 16, which balances memory usage with tracking performance, as later explored in \ref{ssw}.

\begin{figure*}[!h]
    \centering
    \includegraphics[width=\textwidth, trim={0.7cm 0.5cm 0.6cm 0.5cm},clip]{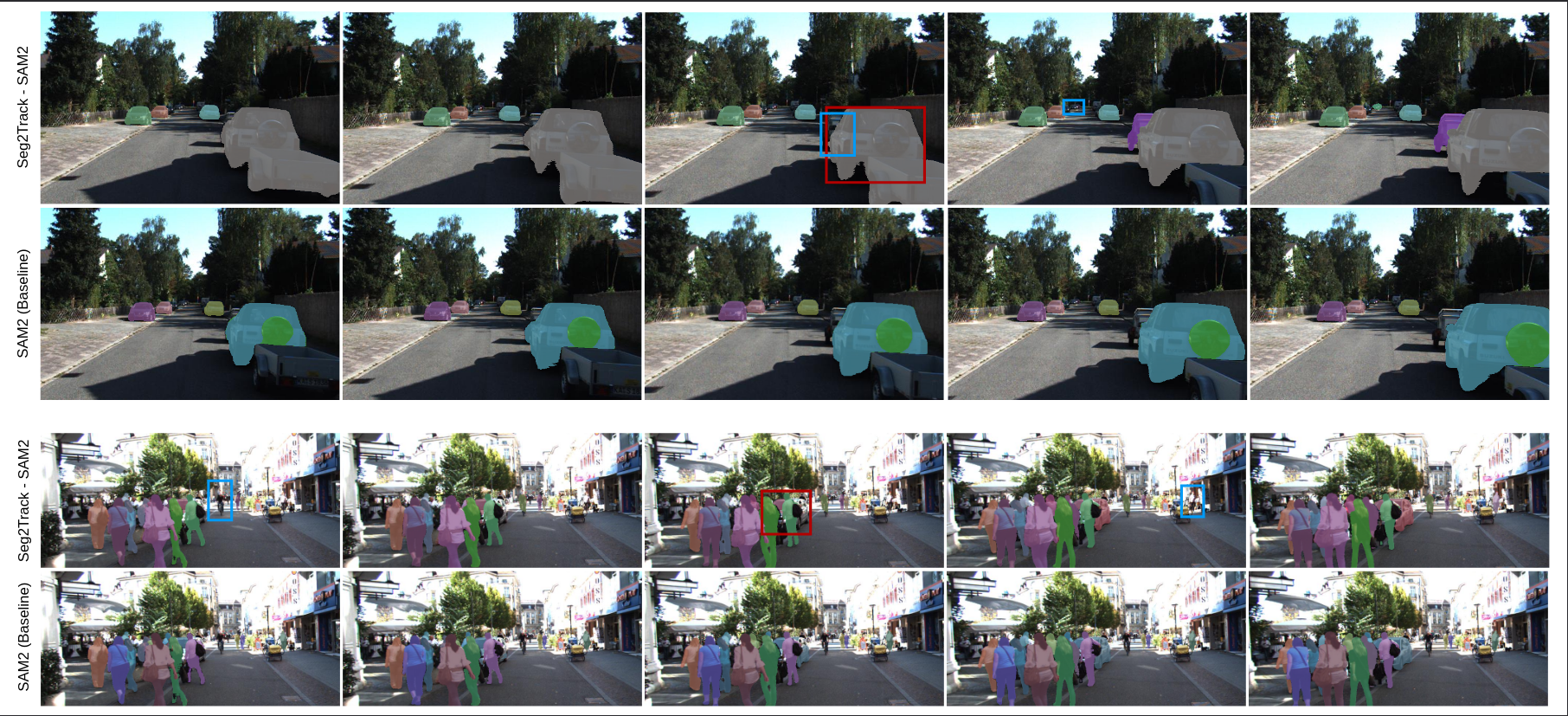}
    \caption{Qualitative comparison of the Seg2Track-SAM2 approach against SAM2 baseline on the MOTS task. The images illustrate the system's ability to manage object tracks in various scenarios. Specifically, blue bounding boxes indicate a track initialization, signifying a newly detected object. In contrast, red bounding boxes represent successful object reinforcement, where a previously tracked object is correctly re-associated with its track.}
    \label{fig:predictions}
\end{figure*}

\subsection{Results}

We evaluate Seg2Track-SAM2 on the KITTI MOTS and MOTS Challenge benchmarks without dataset-specific fine-tuning. Three pretrained object detectors are used to provide a detector-agnostic assessment of the proposed tracking framework. As the detectors are applied without additional training on the target datasets, the contribution of the Seg2Track module can be isolated.

The evaluation is structured to assess three aspects: (i) robustness to detector variation, (ii) the contribution of individual modules to tracking performance, and (iii) the impact of the sliding-window memory strategy on computational efficiency.  We first conduct a detector validation study to analyze the influence of detector choice on overall performance and association accuracy. We then present benchmark results on KITTI MOTS and MOTS Challenge, followed by an ablation study that quantifies the contribution of Track Quality Assessment and Object Association and Filtering. Finally, we evaluate the memory–performance trade-off introduced by the constrained memory strategy.

Detector comparison results are reported in Table~\ref{tab:detector_comparison}. Quantitative benchmark results are presented in Tables~\ref{tab:mots} and~\ref{tab:mots_challenge}, with qualitative examples shown in Fig.~\ref{fig:predictions}. The ablation analysis and memory evaluation are summarized in Tables~\ref{tab:ableton_mots} and~\ref{fig:hack_evaluation}, respectively.

\subsubsection{Detector Evaluation}




To evaluate  the proposed tracking framework across detectors, Seg2Track-SAM2 is tested with multiple pretrained detectors (Table~\ref{tab:detector_comparison}). This study assesses whether the tracking performance and association improvements remain consistent across different detection architectures.

The results indicate a trade-off between detection accuracy (DetA) and association performance (AssA). On KITTI MOTS, YOLOv11 achieves the highest HOTA for both Car (74.63\%) and Pedestrian (51.73\%) classes, driven primarily by higher DetA while maintaining competitive AssA. Although RT\_DETR attains the highest AssA for cars (79.08\%), its lower detection accuracy results in reduced overall HOTA. YOLOv26 achieves comparable performance but does not surpass YOLOv11 in overall tracking accuracy.

On MOTS Challenge, YOLOv26 attains the highest HOTA (65.02\%) and DetA (66.06\%), while YOLOv11 achieves the strongest association performance (AssA = 65.88\%) and remains competitive in overall HOTA (64.09\%). Across both datasets, association performance remains stable across detectors, indicating that the improvements introduced by Seg2Track are not limited to a specific detection backbone.

Considering performance across datasets and object classes, YOLOv11 provides a consistent balance between detection and association performance. It is therefore selected as the default detector in subsequent experiments.

\subsubsection{KITTI MOTS} 

On the KITTI MOTS test set, Seg2Track-SAM2 ranks fourth overall in both object classes, achieving HOTA scores of 74.11\% for cars and 59.93\% for pedestrians. All results are obtained without dataset-specific fine-tuning, relying exclusively on pretrained models.

To isolate the contribution of the proposed Seg2Track module, we compare against a baseline configuration in which SAM2 is prompted only for object initialization, without track refinement or quality assessment. Incorporating Seg2Track yields consistent improvements in both detection and association performance, with gains of approximately 6\,pp in DetA and 6\,pp in AssA. These improvements result from more effective suppression of false-positive propagation and improved management of track continuity over time.

Compared to SOTA methods, Seg2Track-SAM2 achieves the highest association accuracy on KITTI MOTS. As reported in Table~\ref{tab:mots}, AssA improves by 8\,pp for cars and 6\,pp for pedestrians relative to the strongest competing approaches. These gains reflect enhanced identity consistency across frames, attributable to the Track Quality Assessment and Object Association and Filtering modules, which selectively reinforce uncertain tracks and limit the persistence of unreliable masks.

Detection accuracy (DetA) remains lower than that of fully supervised approaches. This behavior is primarily associated with the absence of dataset-specific fine-tuning and the use of a general-purpose detector (YOLOv11), as well as the temporary persistence of spurious detections before removal by the Track Quality Assessment mechanism. This design choice favors training-free deployment and cross-dataset applicability over benchmark-specific optimization.

\subsubsection{MOTS Challenge}

On the MOTS Challenge benchmark, Seg2Track-SAM2 achieves an HOTA score of 61.1\%, ranking fourth overall without dataset-specific fine-tuning (Table~\ref{tab:mots_challenge}).

To isolate the contribution of Seg2Track, we compare against a baseline configuration in which SAM2 performs tracking without track refinement or quality assessment. The integration of Seg2Track improves HOTA by 5.6\,pp and increases association accuracy (AssA) from 58.5\% to 66.0\%, reflecting enhanced identity consistency across frames.

Compared to competing approaches, Seg2Track-SAM2 achieves the highest AssA score (66.0\%) on this benchmark. This improvement is attributed to the Track Quality Assessment and Object Association and Filtering mechanisms, which reduce the persistence of unreliable tracks and reinforce ambiguous associations.

Detection accuracy (DetA = 57.9\%) remains lower than that of fully supervised methods, consistent with observations on KITTI MOTS. This behavior is associated with the use of pretrained detectors without benchmark-specific fine-tuning. Localization accuracy (LocA = 85.0\%) remains competitive and improves substantially over the baseline (68.6\%), indicating stable mask refinement once tracks are established.



\begin{table}[t]
\centering
\caption{Results on the MOTS Challenge benchmark. The superscript (in blue) indicates the column-wise ranking of the methods.}
{\renewcommand{\arraystretch}{1.3}
\resizebox{\columnwidth}{!}{
\begin{tabular}{l|cccc}
\tableline
Model
& HOTA $\uparrow$ 
& DetA $\uparrow$ 
& AssA $\uparrow$ 
& LocA $\uparrow$ \\
\hline

EMNT ~\cite{9841421}
& 66.0{\color{blue}$^1$} 
& 71.0{\color{blue}$^2$} 
& 62.3{\color{blue}$^3$} 
& 86.0{\color{blue}$^2$} \\

ReMOTSv2 ~\cite{YANG2022104514}
& 65.4{\color{blue}$^2$} 
& 71.5{\color{blue}$^1$} 
& 60.9{\color{blue}$^4$} 
& 85.9{\color{blue}$^3$} \\

ReMOTS ~\cite{yang2021remotsselfsupervisedrefiningmultiobject}
& 65.0{\color{blue}$^3$} 
& 71.5{\color{blue}$^1$} 
& 60.2{\color{blue}$^5$} 
& 85.9{\color{blue}$^3$} \\

MPNTrackSeg ~\cite{braso2022multi}
& 57.4{\color{blue}$^5$} 
& 60.8{\color{blue}$^5$} 
& 55.5{\color{blue}$^7$} 
& 83.1{\color{blue}$^6$} \\

OPITrack ~\cite{gullapalli2021opitrack}
& 44.9{\color{blue}$^7$} 
& 64.7{\color{blue}$^4$} 
& 31.8{\color{blue}$^8$} 
& 86.2{\color{blue}$^1$} \\ \hline

\textit{(No fine-tuning w/YOLOv11)} & & & & \\

Baseline (SAM2)
& 55.5{\color{blue}$^6$} 
& 52.6{\color{blue}$^7$} 
& 58.5{\color{blue}$^6$} 
& 68.6{\color{blue}$^7$} \\

Ours (Seg2Track-SAM2)
& 61.1{\color{blue}$^4$} 
& 57.9{\color{blue}$^6$} 
& \textbf{66.0}{\color{blue}$^1$} 
& 85.0{\color{blue}$^5$} \\ \hline

\tableline
\end{tabular}
}
}
\label{tab:mots_challenge}
\end{table}

\subsubsection{Ablation Study} 
We evaluate the contribution of each Seg2Track component on the KITTI MOTS training set using detections from both TrackR-CNN and YOLOv11. Modules are incrementally introduced to quantify the individual and combined effects of Track Quality Assessment (TQA) and Object Association and Filtering (OAF). Results are reported in Table~\ref{tab:ableton_mots}.

Adding TQA to the baseline configuration produces a substantial improvement in tracking performance, with HOTA increasing by approximately 13\,pp for the car class and 10\,pp for the pedestrian class. This improvement is primarily associated with higher association accuracy, as filtering unreliable tracks limits the persistence of incorrect identities over time.

Introducing OAF further improves performance by reinforcing ambiguous tracks and enabling recovery in cases of mask degradation due to appearance variation or partial occlusion. While its individual contribution is smaller than that of TQA, OAF consistently improves both HOTA and AssA across detectors.

Importantly, similar trends are observed when using TrackR-CNN and YOLOv11 detections, indicating that the performance gains are not detector-specific. These results indicate that the improvements in association accuracy and overall tracking performance are attributable to the proposed Seg2Track modules rather than the detection backbone.

\begin{table}[t]
\centering
\caption{Ablation study on the 2D KITTI MOTS training set, for \textit{car} and \textit{pedestrian} classes.}
 {\renewcommand{\arraystretch}{1.22}
\begin{tabular}{ccccc}
\tableline
\rowcolor{white}
Detector & TQA &  OAF & HOTA Car $\uparrow$  & HOTA Ped $\uparrow$  \\
\hline
\textit{(TrackR-CNN)} & & & &  \\
 \checkmark & & & 64.21 & 46.16 \\
 \checkmark & \checkmark & & 75.98 & 55.18 \\
 \checkmark & \checkmark & \checkmark & 77.50  & 55.73 \\
\hline
\textit{(YOLOv11)}  & & & & \\
 \checkmark & & & 63.86  & 50.79 \\
 \checkmark & \checkmark & & 75.36 & 60.69 \\
 \checkmark & \checkmark & \checkmark & 76.95  & 60.86 \\
\tableline
\end{tabular}
}
\label{tab:ableton_mots}
\end{table}

\begin{figure}[t]
    \centering
\includegraphics[width=0.5\textwidth, trim={0.0cm 0.0cm 0.0cm 0.0cm},clip]{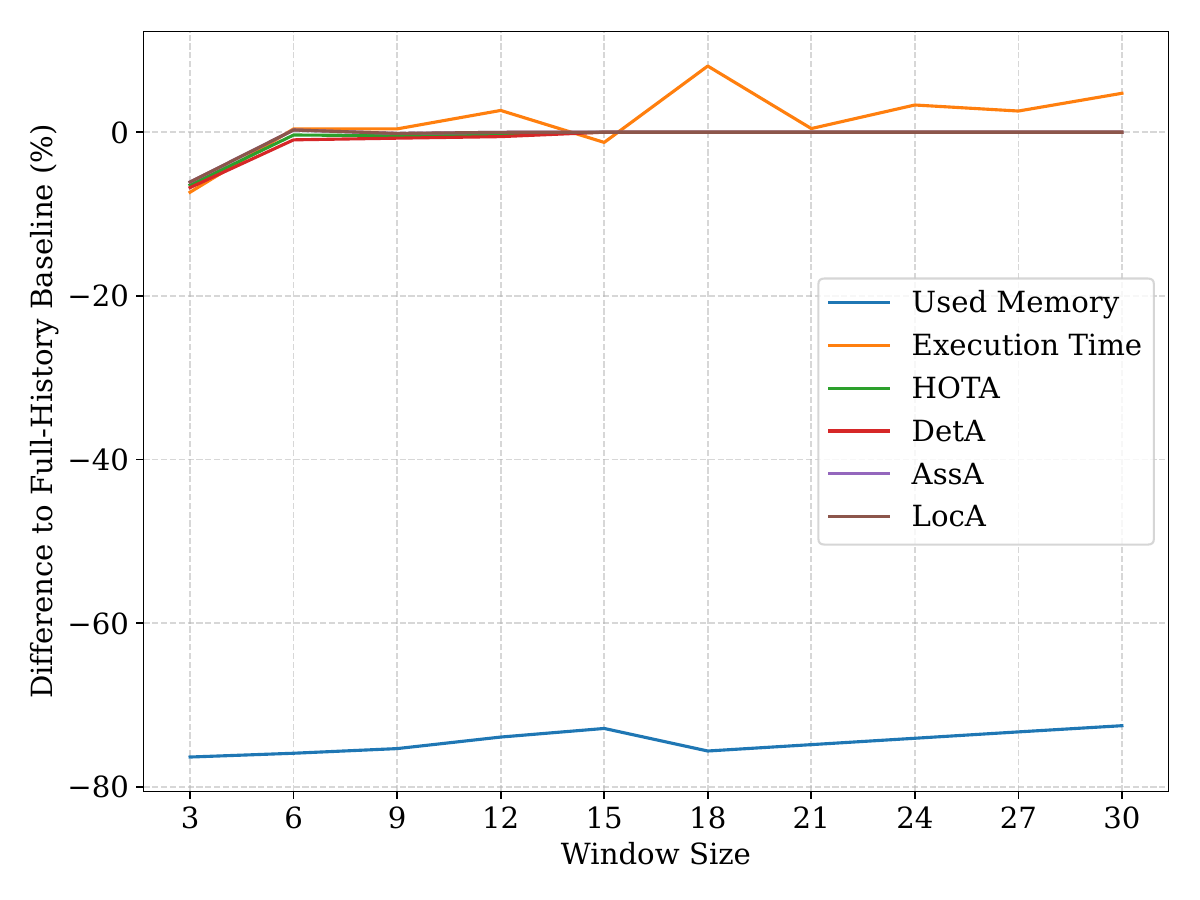}
    \caption{
   Illustration of the effect of varying the backward temporal window size ($T_w$) on Seg2Track-SAM2 performance. The y-axis reports the percentage difference with respect to the full-history baseline without the sliding state window. As the window size increases, HOTA, DetA, AssA, and LocA remain stable, with only minor fluctuations around the baseline. In contrast, memory usage shows a consistent decrease, surpassing 75\% for smaller window sizes compared to the full-history configuration.
}
    \label{fig:hack_evaluation}
\end{figure}

\subsubsection{Sliding State Window}
\label{ssw}

To evaluate the effect of the Sliding State Window, we ran Seg2Track-SAM2 on the KITTI MOTS training set using varying window sizes, from 3 to 30 past states per mask, and recorded total memory usage, execution time and HOTA metrics.

As illustrated in Fig.~\ref{fig:hack_evaluation}, the sliding state window reduces memory usage by about 75\% for smaller window sizes, with memory consumption scaling progressively as the window size grows. In terms of execution time, smaller windows provide only modest speedups, while increasing the window size above 9 introduces a computational overhead due to increased management of past states. HOTA performance metrics stabilize near baseline values at a window size of approximately 15, indicating that most relevant temporal information is retained within this range. This behavior reflects a characteristic of object trajectories in KITTI sequences, where objects rarely remain continuously occluded or untracked beyond a dozen frames, making longer histories redundant.

The results indicate a trade-off between memory efficiency and tracking performance. Selecting a moderate window size substantially reduces memory usage while preserving accuracy, supporting deployment on memory-constrained hardware and real-time systems such as autonomous-driving platforms. Furthermore, the results show that training-free approaches such as Seg2Track-SAM2 can maintain stable tracking performance even when the amount of retained historical state information is limited.

\section{CONCLUSION}
This paper presents Seg2Track-SAM2, a MOTS framework that leverages the generalization capability of SAM2 while addressing limitations in identity preservation and memory efficiency. The framework incorporates dedicated modules for track quality assessment, binary mask generation, and object association to improve temporal consistency and control false-positive propagation.

Evaluation on the KITTI MOTS and MOTS Challenge benchmarks shows competitive performance without dataset-specific fine-tuning, achieving the highest association accuracy among compared methods. The results further indicate improved identity stability under occlusion and appearance variation, as well as reduced memory consumption through a sliding-window state representation.

\section*{ACKNOWLEDGMENT}
This work has been supported by the project PharmaRobot (ref. COMPETE2030-FEDER-01478600), and funded by Fundação para a Ciência e a Tecnologia (FCT), Portugal. This work was also supported by the ISR-UC FCT grant UIDB/00048/2025 (DOI: 10.54499/UIDB/00048/2025)
\bibliographystyle{IEEEtran}
\bibliography{abbrev_short,ref_improved}

\end{document}